\pdfoutput=1

\documentclass[11pt]{article}

\usepackage[]{acl}

\usepackage{times}
\usepackage{latexsym}
\usepackage{amssymb}
\usepackage{dblfloatfix}

\usepackage{graphicx}
\usepackage{subcaption}
\usepackage{adjustbox}
\usepackage[T1]{fontenc}

\usepackage[utf8]{inputenc}
\usepackage{graphicx}
\usepackage{multirow}
\usepackage{microtype}
\usepackage{amsmath}
\usepackage{inconsolata}

\usepackage{booktabs}
\usepackage[compact]{titlesec}

\newcommand{\methodNameNoSpace}{\textsc{DaFair}}
\newcommand{\methodName}{\textsc{DaFair} }
\newcommand{\methodFullName}{Demographic-Agnostic Fairness }
\newcommand{\secondMethodName}{\textsc{Semi-DaFair}}
\usepackage{xcolor}

\title{Leveraging Prototypical Representations for Mitigating Social Bias without Demographic Information}

\author{Shadi Iskander \hspace{2em} Kira Radinsky \hspace{2em} Yonatan Belinkov \\
  {\tt shadi.isk@campus.technion.ac.il} \\
  {\tt kirar@cs.technion.ac.il \hspace{2em} belinkov@technion.ac.il} \\
{Technion -- Israel Institute of Technology}}
\begin{document}
\maketitle
\begin{abstract}
Mitigating social biases typically requires identifying  the social  groups associated with each data sample. In this paper, we present \methodNameNoSpace, a novel approach to address social bias in language models. Unlike traditional methods that rely on explicit demographic labels, our approach does not require   any such information. Instead, we leverage predefined  prototypical demographic texts and incorporate a regularization term during the fine-tuning process to mitigate bias in the model's representations. Our empirical results across two tasks and two models demonstrate the effectiveness of our method compared to previous approaches that do not rely on labeled data. Moreover, with limited demographic-annotated data, our approach outperforms common debiasing approaches.\footnotemark \footnotetext{Our code is available at \url{https://github.com/technion-cs-nlp/DAFair}}
\end{abstract}

\section{Introduction and Background}


The presence of social bias in training data presents a significant challenge in the development of language models for real-world applications. While these models possess remarkable capabilities, biases within the data can lead to unfair outcomes. Mitigating these biases is crucial, but it becomes particularly challenging when acquiring or accessing sensitive attribute labels is costly or unfeasible.  

Studies showed that language models have the ability to capture demographic information about the writer, including race or gender, within their representations \cite{caliskan2017semantics,DBLP:conf/naacl/ZhaoWYOC18}. However, this capability can introduce unintended biases, leading to discriminatory outputs \cite{10.1145/3287560.3287572}. 

\begin{figure}[!t]
\includegraphics[width=1\linewidth]{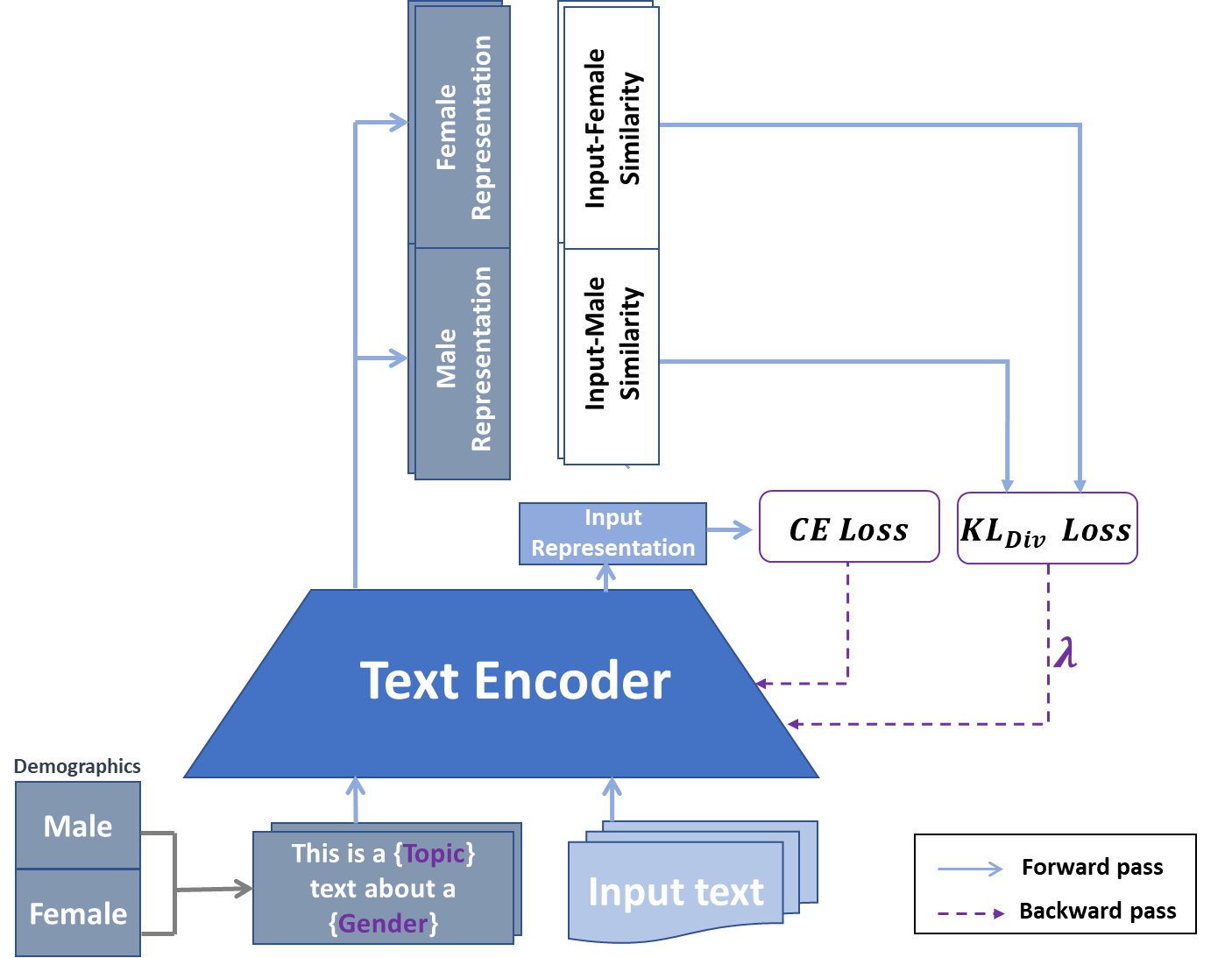}
\vspace{0ex}
\caption{Our debiasing method consists of defining task-specific representations for each social attribute, measuring similarity in the representation space for each example, and utilizing the KL loss to encourage uniform probabilities across social groups.}
  \label{fig:method}
\end{figure}  

Common approaches for social bias mitigation require explicit annotation of biases for each sample in the data \cite{DBLP:journals/corr/BeutelCZC17,zhang2018mitigating}. Recent concept removal methods \cite{DBLP:conf/acl/RavfogelEGTG20,DBLP:conf/icml/RavfogelTGC22,ravfogel-etal-2022-adversarial,iskander-etal-2023-shielded} have shown promise in addressing social bias by removing sensitive attributes. These approaches rely on training classifiers for predicting the sensitive attribute, and training such classifiers typically requires a significant amount of annotated data. 

A promising line of research has emerged that aims to mitigate bias without relying on explicit information about the biases present in the data. For instance,
Just Train Twice (JTT) \cite{liu2021just}  employs a two-step training process. In the second step, a second model is trained on up-weighed training examples that were misclassified by the first model. Another method is
BLIND \cite{DBLP:journals/corr/abs-2212-10563}, which
  introduces a success detector and down-weighs examples for which the detector accurately predicts the outcome. 

In this paper, we propose \textbf{\methodNameNoSpace}: \textbf{D}emographics-\textbf{A}gnostic \textbf{Fair}ness, a novel approach for mitigating social bias during the fine-tuning process of language models, without relying on demographic information. Our approach aims to ensure equal similarity between the representation of a text and prototypical representations of different demographic groups. For instance, when classifying a biographical text of a person into their profession, our method aims to make the representation of the text equally similar to the representations of both males and females. More concretely, \methodName first defines prototypical texts, such as ``This is a biography about a male'' and ``This is a biography about a female''. It then adds a regularization term that makes the representation of a training example equally similar to the representations of each of the prototypical texts (Figure~\ref{fig:method}).

Furthermore, we extend our approach to scenarios where limited demographic-annotated data is available. In such cases, we obtain the prototypical representation by averaging the sample representations corresponding to each social attribute. 

We evaluate the effectiveness of \methodName and its extension on two tasks: occupation prediction and sentiment analysis of twitter posts. In these tasks, we investigate the performance of our approach under the settings of limited demographic labels or no labels at all, reflecting real-world scenarios where labeled data is challenging to obtain.  The experimental results with two base models demonstrate that our approach outperforms previous approaches that  do not rely on demographic information, as well as common approaches with limited data.

\section{Methodology}

Assume a dataset $\smash{D = \{t_i, y_i, z_i\}_{i=1}^n}$ of input texts $\smash{t_i \in \mathcal{T}}$, main task labels $\smash{y_i \in \mathcal{Y}}$, and sensitive attributes $\smash{z_i \in \mathcal{Z}}$ that correspond to discrete demographic attributes, such as race. This sensitive attribute can either be unobserved during training or available in a small subset of the data.  Our aim is to learn a model $\smash{F : \mathcal{T} \rightarrow \mathbb{R}^{|\mathcal{Y}|}}$  that  does not rely on the sensitive attribute $z_i$ in its prediction. 

\subsection{\methodFullName Approach}

 Our method, depicted in Fig.\ \ref{fig:method}, involves several key steps to mitigate social bias. First, we establish multiple representations for each group of sensitive attributes (Section \ref{sec:representations}). During fine-tuning, we measure 
similarity between the representation of an example and each attribute representation. These similarities are then transformed into a probability distribution. Subsequently, we use the Kullback-Leibler (KL) divergence loss \cite{kullback1951information} to compare the predicted probability distribution with a uniform distribution (Section \ref{sec:kl_loss}). This loss term encourages the model to mitigate bias by penalizing deviations from a uniform distribution, promoting fair and unbiased predictions.



\subsubsection{Social Attribute Representations}
\label{sec:representations}
We employ two approaches to define representations for social attribute groups, depending on the availability of labels: no labels, or few labels. 

     \paragraph{Pre-defined Representations (No Labels).} In the absence of labeled data, we leverage semantic similarity and define pairs of texts that capture the models' understanding of text describing different social attribute groups. For example, to represent gender in an occupation prediction task we can use the encoder's representations of ``This biography is about a man'' and ``This biography is about a woman''. To generate these pre-defined representations, we employ a generative model. We provided ChatGPT \cite{chatgpt} with a description of the approach, \methodNameNoSpace, along with a description of each dataset and task, and instructed the model to produce 10 pairs of prototypical texts for each task. The prototypical texts (Tables \ref{table:pd-bios} and \ref{table:pd-moji}) and the full prompt (Figure \ref{fig:prompt}) are provided in the appendix. 
    
     \paragraph{Data-driven Representations (Few Labels).} When a limited number of labels are available, we leverage the representations generated by the text encoder to derive data-driven representations for each labeled group. Specifically, we calculate the mean representation of each labeled group using the available labeled samples. We call this method \textbf{\secondMethodName{}}.
    
    

We will assume a binary case for simplicity and denote the pair of representations as $\smash{[X_{A} , X_{B}]}$.\footnote{Our approach can be extended to handle multiple social attribute groups, denoted as $\smash{[X_A, X_B, X_C, ...]}$.}


\subsubsection{Ensemble of Representations}
Inspired by \citet{stacey-etal-2020-avoiding}, we adopt an ensemble approach by leveraging multiple pairs of representations instead of using a single pair. We denote the ensemble of representations as $\smash{\{[X_{A}^j, X_{B}^j]\}_{j=1}^K}$, where $\smash{K}$ represents the number of pairs.

In the case of pre-defined representations, we use multiple pre-defined pairs that capture different perspectives. For data-driven representations, we divide the labeled data into K partitions and calculate the mean representation for each partition, resulting in K pairs of representations. 

By incorporating an ensemble of representations, we aim to capture a diverse range of information and perspectives related to biases. 

\subsubsection{Calculating KL Loss}
\label{sec:kl_loss}
During fine-tuning, we calculate the similarity between the representation of example $\smash{X_{i}}$ and each pair of attribute representations using dot product:

\begin{equation}
    [sim_{A}^j, sim_{B}^j] = X_{\text{i}} \cdot [X_{A}^j, X_{B}^j] 
\end{equation}
 Then we apply the softmax function $\sigma(a, b) = \frac{e^a}{e^a + e^b}$ to obtain the similarity distribution:
\begin{equation}
    d_{sim}^j = \sigma(sim_{A}^j, sim_{B}^j)
\end{equation}

To calculate the overall KL loss, we compute KL divergence between each of the similarity distributions $\smash{d_{sim}^j}$ and a uniform distribution $\smash{d_{uni}}$:
\begin{equation}
    L_{kl} = \sum_{j=1}^K D_{KL}(d_{sim}^j, d_{uni})
\end{equation}
Finally, we compute the total loss:
\begin{equation}
    L_{total} = L_{ce} + \lambda L_{kl},
\end{equation}
where $\smash{L_{ce}}$ is the usual cross-entropy loss. 
The hyper-parameter $\lambda$  adjusts the balance between task performance and fairness, providing flexibility to prioritize either aspect.

\section{Experimental Setup}
\subsection{Tasks}
\label{sec:tasks}
We conduct experiments on two classification tasks: occupation prediction and sentiment analysis, focusing on social bias related to gender and race.

\paragraph{Occupation Prediction.} We use the Bias in Bios Dataset \cite{10.1145/3287560.3287572}.
The task involves predicting the occupation of individuals based on their biographical information. The dataset consists of 394K biographies of 28 professions, with gender annotations. 

\paragraph{Twitter Sentiment Analysis.} We follow the setup of \citet{elazar-goldberg-2018-adversarial}, who leveraged a Twitter dataset originally gathered by \citet{blodgett-etal-2016-demographic}.  \citet{elazar-goldberg-2018-adversarial} used emojis in the tweets to derive sentiment labels for the classification task. Tweets are labeled with sociolects—African American English (AAE) or Standard American English (SAE)—based on the author's geo-location, serving as a proxy for their racial identity. We work with a subset of 100K samples, consistent with \citet{DBLP:journals/corr/abs-2212-10563}.

\subsection{Models}
We use two pre-trained text encoders: BERT \cite{DBLP:conf/naacl/DevlinCLT19} and DeBERTa-V3 \cite{he2022debertav3}. 
 By considering two diverse tasks and different models, we can evaluate the effectiveness of our approach in mitigating social bias in various contexts and with different model architectures.
\subsection{Metrics}
\paragraph{Performance Evaluation.} We evaluate the model's accuracy (\textbf{Acc}) on the downstream task to ensure that it has not been significantly affected.

\paragraph{Fairness Assessment.} To evaluate extrinsic bias, we align with previous work \cite{10.1145/3287560.3287572,DBLP:conf/acl/RavfogelEGTG20} and use the True Positive Rate Gap (\textbf{TPR-GAP}) as the main fairness metric to assess performance disparities across different protected attribute groups. Following the guidelines in \citet{orgad2022choose} for a comprehensive evaluation, we also incorporate statistical fairness metrics: \textbf{Independence}, \textbf{Separation} and \textbf{Sufficiency}. The metrics details and calculation procedures are provided in Appendix \ref{app:fairness-metrics}.
\renewcommand{\arraystretch}{1.4}
\setlength{\tabcolsep}{10pt}
\begin{table*}[t]
\centering

\begin{tabular}{p{2.5cm}  cc  cc}
\toprule
 & \multicolumn{2}{c}{Occupation Prediction} & \multicolumn{2}{c}{Sentiment Analysis} \\
\cmidrule(lr){2-3} \cmidrule(lr){4-5}
Method & Accuracy $\uparrow$ & TPR-GAP $\downarrow$ & Accuracy $\uparrow$ & TPR-GAP $\downarrow$ \\
\midrule
Original & $83.43\pm{\scriptstyle0.08}$ & $14.66\pm{\scriptstyle0.51}$ &
$79.18\pm{\scriptstyle0.22}$ & $25.34\pm{\scriptstyle0.82}$ \\
\midrule
JTT   & $81.34\pm{\scriptstyle0.81}$ & $14.19\pm{\scriptstyle1.08}$ &
$78.08\pm{\scriptstyle1.21}$ & $23.67\pm{\scriptstyle1.87}$ \\
BLIND & $82.52\pm{\scriptstyle0.23} $ & $13.76\pm{\scriptstyle1.18}$ &
$76.45\pm{\scriptstyle0.67}$ & $22.68\pm{\scriptstyle2.40}$\\
\methodName & $82.32\pm{\scriptstyle0.13}$ & $12.29\pm{\scriptstyle0.32}$ &
$77.20\pm{\scriptstyle1.17}$ & $21.72\pm{\scriptstyle0.82}$ \\
\bottomrule
\end{tabular}
\caption{Evaluation results for occupation prediction and sentiment analysis tasks with BERT as the text encoder.}
\label{table:results-bert}
\end{table*}

 \subsection{Compared Methods}
 \label{sec:compared}
We compare our approach with several methods for bias mitigation and with a baseline (\textbf{Original}) without any debiasing procedure.

We compare with two existing methods that do not rely on demographic information: 

\textbf{JTT} \cite{liu2021just}, which trains in a second phase on up-weighed hard examples.

\textbf{BLIND} \cite{DBLP:journals/corr/abs-2212-10563}, which uses a success detector to down-weigh biased examples.


\label{app:comparing_methods}
When only limited demographic labeled samples are available, we evaluate three methods:

\textbf{INLP} \cite{DBLP:conf/acl/RavfogelEGTG20} removes linear information from the neural
representation by iteratively training a linear
classifier to predict the demographic attribute from the representation,
then projecting the representations
to the null-space of the linear classifier.

\textbf{RLACE} \cite{ravfogel-etal-2022-adversarial} is similar to INLP with the goal of linear information removal from the neural representations. However, it uses a different approach of a linear minimax game.

\textbf{IGBP} \cite{iskander-etal-2023-shielded} overcome the drawbacks of INLP and RLACE which only remove linearly encoded information, and removes non-linear information from representations by gradient-based projections.

\subsection{Settings}
\paragraph{No Demographic Labels.} In this setting, we explore scenarios where demographic labels are not available. We evaluate the performance of demographic-agnostic methods: JTT, BLIND and DAFAIR.
\paragraph{Limited Demographic Labels.} Additionally, we investigate a scenario where we have limited access to demographic labels. In this setting, we apply information removal methods along with \secondMethodName while varying the size of the available demographic-labeled data to analyze their effectiveness.

 We run each method using 5 random seeds and report the mean and standard deviation of the test results. More details on training setup and evaluation procedures are described in Appendix \ref{app:setup}.

\subsection{\methodName Hyperparameters}
\label{sec:hyperparam}
Under the setting of no demographic labels, there is no validation set to optimize the selection of prototypical texts or the number of pairs. To avoid dependency on the choice of prototypical representations, we first generate $N>K$ pairs, and within each iteration, we randomly sample $K$ pairs. For all experiments, we set $N=10$ , $K=4$ to capture diverse associations of the training samples with demographic attributes, without relying on an extensive set of pairs. In Section \ref{app:k-effect}, we analyze the impact of $K$ on the model's performance and assess its implications on fairness and bias mitigation.
\paragraph{$\lambda$ Tuning.} To perform $\lambda$ tuning without the need for a validation set with demographic annotations, we adopt \citet{DBLP:journals/corr/abs-2212-10563}'s strategy that prioritizes selecting the most radical parameter, while ensuring that the downstream task accuracy remains above 0.97 of the original accuracy. More details are described in Appendix \ref{app:setup}.

\section{Results and Analysis}
\label{sec:results}
\subsection{No Demographic Labels}
Table \ref{table:results-bert} presents the evaluation results on the occupation prediction and twitter sentiemnt tasks 
using BERT as encoder. In both tasks, Our proposed method, \methodNameNoSpace, achieves a slightly lower accuracy compared to the finetuned model. However, it significantly reduces the TPR-GAP, outperforming BLIND and JTT in mitigating bias related to gender or race. This could be attributed to the fact that JTT and BLIND do not directly address social bias, but assign different weights to examples based on their difficulty. In contrast, \methodName uses a regularization term designed to lower the association of text representation with specific social groups, which might explain the superior reduction of social bias measures. Evaluation with other statistical fairness metrics reveals similar patterns to TPR-GAP (Appendix \ref{app:full-results}). Results with the DeBERTA-V3 model exhibit same trend, as presented in Appendix \ref{app:full-results}.





\begin{figure*}[!t]
\centering
\begin{subfigure}[b]{0.48\linewidth}
\includegraphics[width=\linewidth]{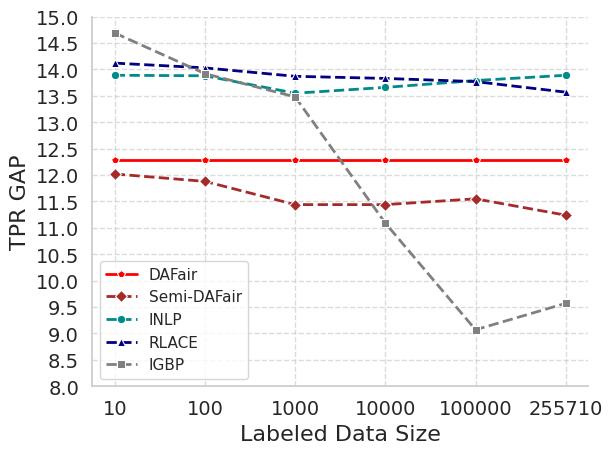}
\caption{Occupation Prediction}
\label{fig:bios_tpr_sizes}
\end{subfigure}%
\begin{subfigure}[b]{0.48\linewidth}
\includegraphics[width=\linewidth]{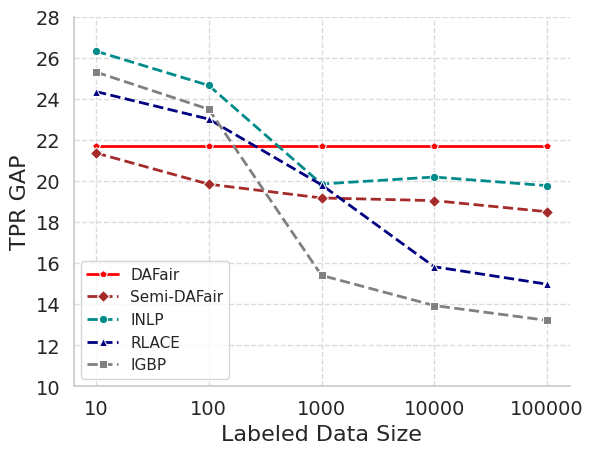}
\caption{Sentiment Analysis}
\label{fig:dial_tpr_size}
\end{subfigure}
\caption{Effect of bias mitigation methods on TPR-GAP with varying labeled data sizes. In scenarios with limited demographic-annotated data, our approach outperforms common debiasing approaches.}
\label{fig:bert_size}
\end{figure*}



\begin{table*}[!b]
\renewcommand{\arraystretch}{1.2}
\setlength{\tabcolsep}{8pt}
\centering

\begin{tabular}{l  cc  cc}
\toprule
 & \multicolumn{2}{c}{Occupation Prediction} & \multicolumn{2}{c}{Sentiment Analysis} \\
\cmidrule(lr){2-3} \cmidrule(lr){4-5}
\centering K & Accuracy $\uparrow$ & TPR-GAP $\downarrow$ & Accuracy $\uparrow$ & TPR-GAP $\downarrow$ \\
\midrule
\centering Original & $83.43\pm{\scriptstyle0.08}$ & $14.66\pm{\scriptstyle0.51}$ &
$79.18\pm{\scriptstyle0.22}$ & $25.34\pm{\scriptstyle0.82}$ \\
\centering 1 & $82.85\pm{\scriptstyle0.18}$ & $13.11\pm{\scriptstyle0.62}$ & $77.52\pm{\scriptstyle1.23}$ & $22.69\pm{\scriptstyle1.81}$ \\
\centering 2 & $82.55\pm{\scriptstyle0.23}$ & $12.84\pm{\scriptstyle0.59}$ & $77.16\pm{\scriptstyle1.12}$ & $22.21\pm{\scriptstyle1.08}$ \\
\centering 4 & $82.32\pm{\scriptstyle0.13}$ & $12.29\pm{\scriptstyle0.32}$ & $77.20\pm{\scriptstyle1.17}$ & $21.72\pm{\scriptstyle0.82}$ \\
\centering 8 & $82.29\pm{\scriptstyle0.30}$ & $12.20\pm{\scriptstyle0.25}$ & $77.47\pm{\scriptstyle0.94}$ & $22.17\pm{\scriptstyle1.17}$ \\
\bottomrule
\end{tabular}
\caption{Effect of varying K on accuracy and TPR-GAP for occupation prediction and sentiment analysis tasks.}
\label{table:varying-k}
\end{table*}

\subsection{Limited Demographic Labels}

\label{sec:few-demo}
Figure \ref{fig:bert_size} presents TPR-GAP for both tasks under different levels of labeled data for social attributes,  
showcasing the performance of various debiasing methods (Section \ref{sec:compared}), including our proposed methods, \methodName and \secondMethodName.\footnotemark \footnotetext{Accuracy figures are available in Appendix \ref{app:acc}.}
While using no labels (horizontal solid lines), \methodName outperforms other methods even when they are provided a limited number of labels (up to 100 in twitter sentiment and 1000 in occupation prediction). \methodName further benefits from labels (\secondMethodName{} lines), even outperforming prior methods with limited labeled data. With an abundance of labeled data (1000 in  sentiment and 100K in occupation prediction), other methods perform better.

The aim of information removal methods is to learn and neutralize decision boundaries between different social attributes using supervised learning. However, limited labeled examples may hinder classifiers from modeling social attribute subspace in high-dimensional spaces. In \secondMethodName, we aim to mitigate associations with specific social groups. Surprisingly, a small set of labeled data seems sufficient for this purpose, as more labeled data does not offer additional benefits.


\subsection{Effect of Number of Prototypical Texts}
\label{app:k-effect}
To investigate the effect of the number of prototypical text pairs (K) on model performance, we conducted experiments with varying K values of (1, 2, 4, 8). The results presented in Table \ref{table:varying-k} reveal that all K values contributes to the reduction of the TPR-GAP without affecting accuracy. While larger values of K result in more substantial reductions, the incremental improvements become less significant for $K > 2$. These findings suggest that a small K may be sufficient for \methodNameNoSpace.

\section{Conclusion}
We introduced \methodNameNoSpace, a novel approach for mitigating social bias in language models without explicit demographic information. Our method leverages semantic similarity to manipulate the model's text representations during finetuning to promote fairness. Experimental results on two tasks and under different settings demonstrated the effectiveness of \methodName in reducing bias and improving fairness while maintaining competitive downstream task performance, even with limited or no labeled demographic data. With its focus on social bias, \methodName offers a flexible framework adaptable to address other forms of bias through the modification of prototypical texts.

In conclusion, our approach offers a practical and flexible solution for bias mitigation in real-world applications, contributing to the development of fairer language models. 


\section*{Limitations}
While our approach shows promise in mitigating social bias in language models without relying on demographic labels, it is important to recognize its limitations. First, our method relies on predefined texts that represent different social attribute groups, which may not fully capture the complexity and diversity of these attributes. Language models are complex systems, and they may still exhibit bias or unintended associations despite our efforts.

Moreover, it is important to acknowledge that gender is non-binary, and the expirements we conducted were focused on addressing binary gender biases.
Additionally, our analysis of racial biases is centered around the African-American race, using sociolect as a proxy which might be inaccurate. We believe there is a need for more comprehensive research to address biases related to African American rase and other racial and ethnic groups, in a more precise manner.

\section*{Ethics Statement}
The development and implementation of our method for mitigating bias in language models require careful ethical considerations. By employing the KL loss regularization term with non-uniform probabilities, there is a possibility of inadvertently amplifying biases or introducing unintended consequences. 
Additionally, while our aim is to mitigate bias without relying on demographic labels, we acknowledge the need for evaluation and validation to minimize any unforeseen biases that may persist.
To mitigate these risks, we strongly recommend the collection of a small validation set to assess the performance of the system and ensure its alignment with ethical considerations. 
\section*{Acknowledgements}
This research was supported by the Israel Science Foundation (grant 448/20), an Azrieli Foundation Early Career Faculty Fellowship, and an AI Alignment grant from Open Philanthropy. 

\bibliography{anthology,custom}

\clearpage

\appendix






\section{Setup}
\label{app:setup}

\subsection{Training Setup}
\label{app:training}

We conduct experiments using two  models: BERT \cite{DBLP:conf/naacl/DevlinCLT19} and  DeBERTa-V3 \cite{he2022debertav3}. For BERT we used the $\texttt{bert-base-uncased}$ and for DeBERTa we used the $\texttt{microsoft/deberta-v3-base}$, both from the Huggingface library \cite{wolf-etal-2020-transformers}.
We utilize the transformer models as a text encoder, where the input text is transformed into a contextualized representation. The [CLS] token of the encoder is then passed through a linear classifier for the downstream task.
We used a $65$/$10$/$25$ training-validation-test split ratio for all tasks. Training was done with a learning rate of $5e-5$ and a  stochastic gradient descent optimizer for 1 epoch.
\subsubsection{\methodName} To maintain suitable representations in the embedding space, we processed the pre-defined text (Sec \ref{sec:representations}) through the text encoder every 200 batches during fine-tuning. This fixed frequency was found to be effective for stable training and did not require further tuning.
\paragraph{$\lambda$ Tuning} To determine the appropriate value for the parameter $\lambda$, we adopted the approach outlined by \citet{ganin2016domain}. The parameter $\lambda$ was initially set to 0 and gradually adjusted towards a predefined threshold value, denoted as $\lambda_{threshold}$, using a specific schedule.
The schedule for updating $\lambda$ is determined by the following formula:
\begin{equation}
\lambda = (\frac{2}{{1 + \exp(-\gamma \cdot p)}} - 1)\cdot\lambda_{threshold}, 
\end{equation} 
where $p$ represents a measure of progress, and $\gamma$ controls the rate of change. The parameter $\gamma$ allows us to control the speed at which $\lambda$ approaches $\lambda_{threshold}$.

In our experiments, we set $\gamma$ to a fixed value of 5. This schedule allows for stable training in the early stages of training.
In order to optimize the performance of \methodName and \secondMethodName, we conducted a grid search to find the optimal value for the parameter $\lambda_{threshold}$. The search was performed over the following values $\lambda_{threshold} \in (0.1, 0.2, 0.5, 1, 2, 5, 10, 20, 50, 100)$.

\subsubsection{Compared Methods}
\label{app:comparing_methods}
For \textbf{BLIND}, we used their implementation and ran a grid search for $\gamma \in (1, 2, 4, 8, 16)$ and $ T \in (2,4)$.

For \textbf{JTT}, we provide our own implementation and tune  for $\lambda_{up} \in (1, 2, 4, 6, 8, 10)$.

For the \textbf{INLP}, \textbf{RLACE}, and \textbf{IGBP} methods, we utilized the implementations provided by the respective authors. To ensure optimal performance, we conducted hyperparameter tuning specifically for the number of iterations. We applied these post-hoc methods on the model's representations extracted from a fine-tuned model (\textbf{Original}).
In experiments with  a limited number of labeled data for social attributes, we modify the concept removal methods by training the debiasing classifiers on the available labeled data.

\subsection{Evaluation Setup} 
\label{app:eval}

As discussed in Section \ref{sec:hyperparam}, to perform parameter tuning without the need for a validation set with demographic annotations, we adopt \citet{DBLP:journals/corr/abs-2212-10563}'s strategy that prioritizes selecting the most radical hyperparameters, while ensuring that the downstream task accuracy remains above 0.97 of the original accuracy.  This is done for all methods.

Intuitively, increasing the regularization term weight (in our case: $ \lambda_{threshold}$) promotes fairness by encouraging the model to distribute its predictions more evenly among different social groups. However, it can also lead to a decrease in task accuracy if applied excessively. By setting a threshold of 0.97 for the accuracy, we strike a balance between bias mitigation and maintaining competitive performance.


\section{Fairness Metrics}
\label{app:fairness-metrics}
\paragraph{TPR-GAP} We calculate the True Positive Rate (TPR) by:

\begin{equation}
\mathrm{TPR_{z,y} = P(\hat{Y}=y|Z=z,Y=y)}
\end{equation}
where $\hat{Y}$ represents the predicted label, $Z$ denotes the protected attribute, and $Y$ represents the true label. 

The TPR Gap is computed as:

\begin{equation}
\mathrm{GAP_{TPR}^{z,y} = TPR_{z,y} - TPR_{z',y}}
\end{equation}
where $z$ and $z'$ correspond to different values of the protected attribute.

To assign a single bias measure across all values of $y$, we calculate the root mean square $\mathrm{GAP_{TPR}^{z}}$.

\begin{equation}
\mathrm{GAP_{TPR}^{z}} = \sqrt{\frac{1}{|C|} \sum_{y=1}^{N} \left(\mathrm{GAP_{TPR}^{z,y}}\right)^2}
\end{equation}
where $C$ represents the total number of label categories.
\paragraph{Statistical Measures.}

Another family of fairness metrics involves statistical measures based on probability distributions. We utilize three key metrics:

\begin{description}
  \item[Independence] This metric measures the statistical dependence between the model's prediction and protected attributes. It employs the Kullback–Leibler divergence between two distributions, namely \(KL(P(y), P(\hat{y}|z = z))\), for \(z\) $\in$ \(Z\). The sum over \(z\) yields a single value describing the model's independence. It assesses how the model's behavior varies across different demographics.

  \item[Separation] This metric assesses the statistical dependence between the model's prediction given the target label and the protected attributes. It utilizes \(KL(P(\hat{y}|y = y), P(\hat{y}|y = y, z = z))\) for all \(y\) $\in$ \(Y\), \(z\) $\in$ \(Z\). This metric is similar to True Positive Rate (TPR) and False Positive Rate (FPR) gaps. It evaluates if the model behaves differently across classes and demographics.

  \item[Sufficiency] This metric measures the statistical dependence between the target label given the model's prediction and the protected attributes. It employs \(KL(P(y|\hat{y} = \hat{y}), P(y|\hat{y} = \hat{y}, z = z))\) for  \(\hat{y}\) $\in$ \(Y\), \(z\) $\in$ \(Z\). The sum over \(\hat{y}\) and \(z\) results in a single value. It intuitively assesses whether a model disproportionately promotes or penalizes specific demographic groups.
\end{description}
To measure these statistical fairness metrics, we used
the AllenNLP fairness library. (\url{https://github.com/allenai/allennlp}).
\section{Full Results}
\label{app:full-results}
Tables \ref{table:full-results-occupation} and \ref{table:full-results-sentiment} present comprehensive results for the occupation prediction and sentiment analysis tasks, respectively, employing BERT as the text encoder. Each method's performance is evaluated across multiple metrics, including Accuracy, TPR-GAP, Independence, Separation, and Sufficiency (Section \ref{app:fairness-metrics}). Here we also see that our proposed method reduces Independence, Separation, and Sufficiency values, in both tasks.

\paragraph{DeBERTA-V3 Results} The evaluation results for the occupation prediction and sentiment analysis tasks using the DeBERTa-V3 model are presented in Tables \ref{table:full-results-occupation-deberta}, and \ref{table:full-results-sentiment-deberta}. In the two tasks, Both JTT and BLIND methods demonstrate some success in reducing bias, although not substantial. However, \methodName, outperforms both JTT and BLIND in terms of mitigating bias related to social attributes. It achieves a lower TPR-GAP, Independence, Separation and Sufficiency in most cases, while maintaining a comparable level of accuracy. This indicates that our approach is more effective in reducing bias without sacrificing the overall performance of the model also with DeBERTa-V3 model.

\renewcommand{\arraystretch}{1.3}
\setlength{\tabcolsep}{10pt}
\begin{table*}[!b]
\centering

\begin{tabular}{p{2.5cm}  ccccc}
\toprule
 & \multicolumn{5}{c}{Occupation Prediction} \\
\cmidrule(lr){2-6}
Method & Accuracy $\uparrow$ & TPR-GAP $\downarrow$ & Indep $\downarrow$ & Sep $\downarrow$ & Suff $\downarrow$ \\
\midrule
Original & $83.43\pm{\scriptstyle0.08}$ & $14.66\pm{\scriptstyle0.51}$ & $0.16\pm{\scriptstyle0.002}$ & $2.59\pm{\scriptstyle0.07}$ & $2.46\pm{\scriptstyle0.08}$ \\
\midrule
JTT   & $81.34\pm{\scriptstyle0.81}$ & $14.19\pm{\scriptstyle1.08}$ & $0.16\pm{\scriptstyle0.002}$ & $2.77\pm{\scriptstyle0.01}$ & $2.22\pm{\scriptstyle0.14}$ \\
BLIND & $82.52\pm{\scriptstyle0.23} $ & $13.76\pm{\scriptstyle1.18}$ & $0.16\pm{\scriptstyle0.002}$ & $2.35\pm{\scriptstyle0.10}$ & $2.21\pm{\scriptstyle0.09}$ \\
\methodName & $82.32\pm{\scriptstyle0.13}$ & $12.29\pm{\scriptstyle0.32}$ & $0.14\pm{\scriptstyle0.001}$ & $1.90\pm{\scriptstyle0.09}$ & $2.20\pm{\scriptstyle0.10}$ \\
\bottomrule
\end{tabular}
\caption{Full results for the occupation prediction task with BERT as the text encoder.}
\label{table:full-results-occupation}
\end{table*}

\begin{table*}[!b]
\renewcommand{\arraystretch}{1.3}
\setlength{\tabcolsep}{10pt}
\centering

\begin{tabular}{p{2.5cm}  ccccc}
\toprule
 & \multicolumn{5}{c}{Sentiment Analysis} \\
\cmidrule(lr){2-6}
Method & Accuracy $\uparrow$ & TPR-GAP $\downarrow$ & Indep $\downarrow$ & Sep $\downarrow$ & Suff $\downarrow$ \\
\midrule
Original & $79.18\pm{\scriptstyle0.22}$ & $25.34\pm{\scriptstyle0.82}$ & $0.17\pm{\scriptstyle0.003}$ & $0.11\pm{\scriptstyle0.003}$ & $0.08\pm{\scriptstyle0.004}$ \\
\midrule
JTT   & $78.08\pm{\scriptstyle1.21}$ & $23.67\pm{\scriptstyle1.87}$ & $0.18\pm{\scriptstyle0.01}$ & $0.11\pm{\scriptstyle0.008}$ & $0.08\pm{\scriptstyle0.006}$ \\
BLIND & $76.45\pm{\scriptstyle0.67}$ & $22.68\pm{\scriptstyle2.40}$ & $0.14\pm{\scriptstyle0.001}$ & $0.05\pm{\scriptstyle0.001}$ & $0.06\pm{\scriptstyle0.001}$ \\
\methodName & $77.20\pm{\scriptstyle1.17}$ & $21.72\pm{\scriptstyle0.82}$ & $0.15\pm{\scriptstyle0.002}$ & $0.09\pm{\scriptstyle0.002}$ & $0.08\pm{\scriptstyle0.004}$ \\

\bottomrule
\end{tabular}
\caption{Full results for the sentiment analysis task with BERT as the text encoder.}
\label{table:full-results-sentiment}
\end{table*}

\begin{table*}[!b]
\renewcommand{\arraystretch}{1.3}
\setlength{\tabcolsep}{10pt}
\centering

\begin{tabular}{p{2.5cm}  ccccc}
\toprule
 & \multicolumn{5}{c}{Occupation Prediction} \\
\cmidrule(lr){2-6}
Method & Accuracy $\uparrow$ & TPR-GAP $\downarrow$ & Indep $\downarrow$ & Sep $\downarrow$ & Suff $\downarrow$ \\
\midrule
Original & $83.42\pm{\scriptstyle0.26}$ & $14.60\pm{\scriptstyle0.81}$ & $0.17\pm{\scriptstyle0.003}$ & $2.67\pm{\scriptstyle0.10}$ & $2.51\pm{\scriptstyle0.09}$ \\

\midrule
JTT & $81.98\pm{\scriptstyle1.70}$ & $14.30\pm{\scriptstyle0.55}$ & $0.18\pm{\scriptstyle0.004}$ & $2.65\pm{\scriptstyle0.04}$ & $2.38\pm{\scriptstyle0.12}$ \\

BLIND & $82.41\pm{\scriptstyle0.55}$ & $13.86\pm{\scriptstyle1.49}$ & $0.16\pm{\scriptstyle0.003}$ & $2.31\pm{\scriptstyle0.12}$ & $2.29\pm{\scriptstyle0.10}$ \\

\methodName & $82.15\pm{\scriptstyle0.32}$ & $12.93\pm{\scriptstyle0.39}$ & $0.14\pm{\scriptstyle0.001}$ & $2.06\pm{\scriptstyle0.09}$ & $2.25\pm{\scriptstyle0.11}$ \\
\bottomrule
\end{tabular}
\caption{Full results for the occupation prediction task with DeBERTa as the text encoder.}
\label{table:full-results-occupation-deberta}
\end{table*}

\begin{table*}[!b]
\renewcommand{\arraystretch}{1.3}
\setlength{\tabcolsep}{10pt}
\centering

\begin{tabular}{p{2.5cm}  ccccc}
\toprule
 & \multicolumn{5}{c}{Sentiment Analysis} \\
\cmidrule(lr){2-6}
Method & Accuracy $\uparrow$ & TPR-GAP $\downarrow$ & Indep $\downarrow$ & Sep $\downarrow$ & Suff $\downarrow$ \\
\midrule
Original & $78.36\pm{\scriptstyle0.93}$ & $29.62\pm{\scriptstyle1.58}$ & $0.19\pm{\scriptstyle0.003}$ & $0.12\pm{\scriptstyle0.004}$ & $0.09\pm{\scriptstyle0.005}$ \\
\midrule
JTT & $77.41\pm{\scriptstyle1.76}$ & $28.06\pm{\scriptstyle2.03}$ & $0.18\pm{\scriptstyle0.013}$ & $0.13\pm{\scriptstyle0.009}$ & $0.08\pm{\scriptstyle0.005}$ \\
BLIND & $77.20\pm{\scriptstyle1.22}$ & $27.32\pm{\scriptstyle2.84}$ & $0.15\pm{\scriptstyle0.001}$ & $0.07\pm{\scriptstyle0.001}$ & $0.08\pm{\scriptstyle0.001}$ \\
\methodName & $77.92\pm{\scriptstyle1.21}$ & $26.24\pm{\scriptstyle1.61}$ & $0.16\pm{\scriptstyle0.002}$ & $0.08\pm{\scriptstyle0.003}$ & $0.07\pm{\scriptstyle0.003}$ \\
\bottomrule
\end{tabular}
\caption{Full results for the sentiment analysis task with DeBERTa as the text encoder.}
\label{table:full-results-sentiment-deberta}
\end{table*}

\subsection{Accuracy Figures}
\begin{figure}[!b]
\centering
\begin{subfigure}[b]{\linewidth}
\includegraphics[width=\linewidth]{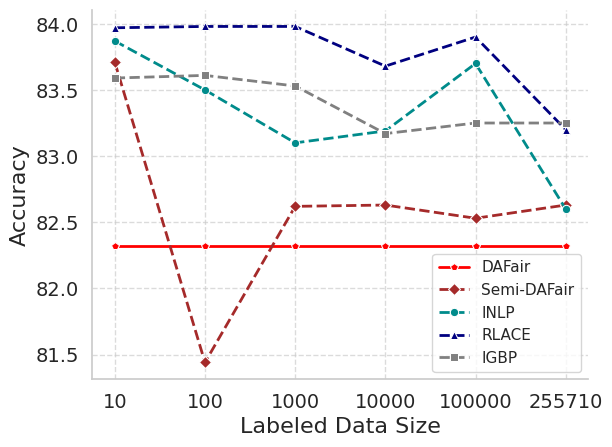}
\label{fig:acc_sizes_bios}
\caption{Occupation Prediction}
\end{subfigure}%
 
\begin{subfigure}[b]{\linewidth}
\includegraphics[width=\linewidth]{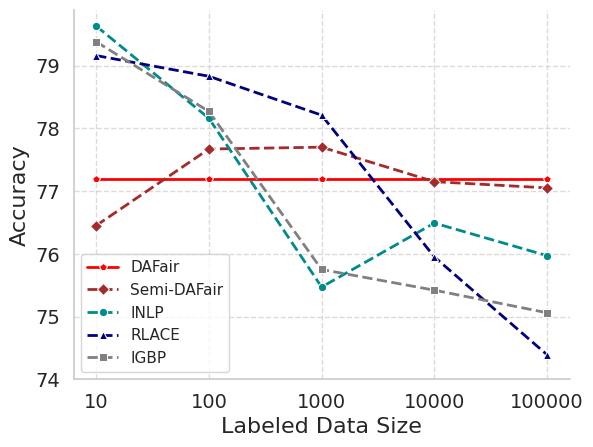}
\caption{Sentiment Analysis}
\label{fig:acc_sizes_dial}
\end{subfigure}
\caption{Effect of bias mitigation methods on accuracy with varying labeled data sizes.}
\label{fig:bert_size_acc}
\end{figure}

In Figure \ref{fig:bert_size_acc} we provide the accuracy of different methods across varying dataset sizes.
\label{app:acc}





\renewcommand{\arraystretch}{1.3}
\setlength{\tabcolsep}{10pt}
\begin{table*}
 \centering
 
    \label{table:pd-moji}
\label{table:pd-bios}
\begin{tabular}{|p{0.45\linewidth}|p{0.45\linewidth}|}
\hline
\textbf{Male Prototypical Texts} & \textbf{Female Prototypical Texts} \\
\hline
This is a biography about a male. & This is a biography about a female. \\
\hline
A man who excelled in his field. & A woman who excelled in her field. \\ \hline
He is known for his achievements in various industries. & She is known for her achievements in various industries. \\ \hline
A prominent male figure in history. & A prominent female figure in history. \\ \hline
His career and accomplishments are well-regarded. & Her career and accomplishments are well-regarded. \\ \hline
This biography focuses on the life of a distinguished man. & This biography focuses on the life of a distinguished woman. \\ \hline
An influential male individual. & An influential female individual. \\
\hline
He made significant contributions to his profession. & She made significant contributions to her profession. \\
\hline
This is a story about a man who shaped his industry. & This is a story about a woman who shaped her industry. \\
\hline
His impact on his field is noteworthy. & Her impact on her field is noteworthy. \\
\hline

\end{tabular}

  \caption{Pre-defined Representations for Male and Female Biographical Texts}
  \label{table:pd-bios}
\end{table*}
\renewcommand{\arraystretch}{1.3}
\setlength{\tabcolsep}{10pt}
\begin{table*}

 \centering
 
\begin{tabular}{|p{0.45\linewidth}|p{0.45\linewidth}|}
\hline
\textbf{AAE Prototypical Texts} & \textbf{SAE Prototypical Texts} \\
\hline
This tweet reflects a [sentiment] from a white writer. & This tweet reflects a [sentiment] sentiment from a black writer. \\
\hline
A tweet expressing a [sentiment] moment by a white individual. & A tweet expressing a [sentiment] moment by a black individual. \\
\hline
A [sentiment] viewpoint shared by a writer using Standart American English. & A [sentiment] viewpoint shared by a writer using African American English. \\
\hline
This post, written in standard English, conveys [sentiment] from a white perspective. & This post, written in AAE, conveys [sentiment] from a black perspective. \\
\hline
A message filled with [sentiment] from a white communicator. & A message filled with [sentiment] from a black communicator. \\
\hline
A white person shares their [sentiment] thoughts in this tweet. & A black person shares their [sentiment] thoughts in this tweet. \\
\hline
This is an example of a tweet with [sentiment] in white sociolect. & This is an example of a tweet with [sentiment] sentiment in AAE. \\
\hline
A tweet written by a white speaker that conveys [sentiment]. & A tweet written by a black speaker that conveys [sentiment]. \\
\hline
This post by a white individual radiates [sentiment] and [sentiment]. & This post by a black individual radiates [sentiment] and [sentiment]. \\
\hline
A [sentiment] perspective presented by a writer using white sociolect. & A [sentiment] perspective presented by a writer using African American English. \\

\hline

\end{tabular}

  \caption{Pre-defined Representations for AAE and SAE Tweet Texts}
    \label{table:pd-moji}
\end{table*}

\clearpage
\begin{figure*}[!b]
\centering
\includegraphics[width=0.8\linewidth,height=0.9\linewidth]{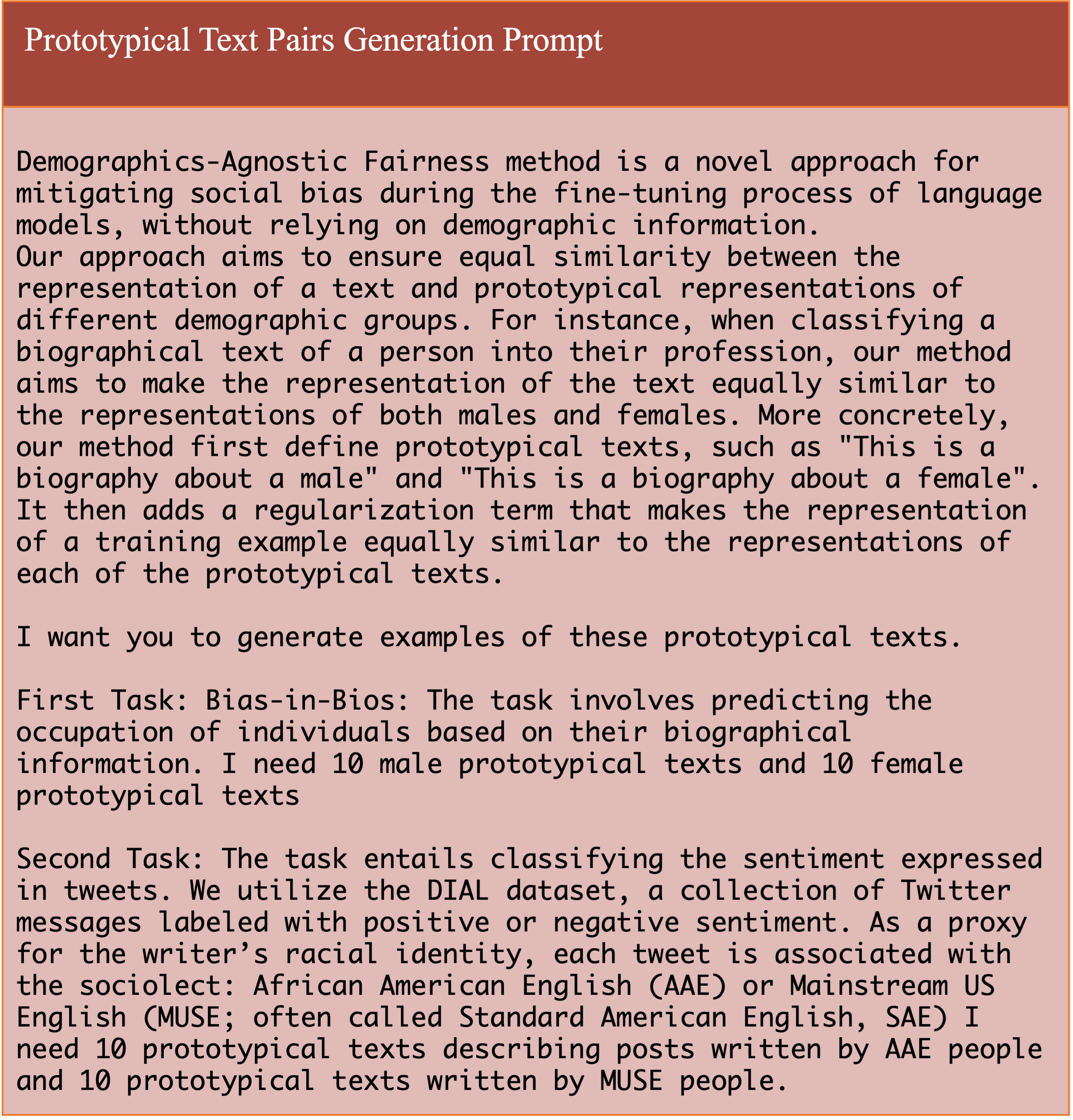}
\caption{The prompt for generating prototypical text pairs.}
\label{fig:prompt}
\end{figure*}
\end{document}